\documentclass[final]{dynafront2026}
\usepackage{booktabs}
\usepackage{microtype}
\usepackage{fontawesome5}

\makeatletter\@secpenalty=0\makeatother 

\title[Finite-Rate Selection in Annealed Gibbs Sampling]{A Finite-Rate Selection Law for Annealed Gibbs Sampling}

\optauthor{%
  \Name{Izhar Ali} \Email{aliz@rowan.edu}\\
  \addr Department of Computer Science, Rowan University
  \AND
  \Name{Muhammad Haris Khan} \Email{dcg967@ku.dk}\\
  \addr Department of Computer Science, University of Copenhagen
  \AND
  \Name{Saeed Ahmad} \Email{saeedahmad@iklab.ai}\\
  \addr IKLAB Inc.
}
\hypersetup{pdftitle={A Finite-Rate Selection Law for Annealed Gibbs Sampling},pdfauthor={Izhar Ali, Muhammad Haris Khan, Saeed Ahmad}}

\newcommand{\qeq}{q_{\mathrm{eq}}}
\newcommand{\qchain}{q_{\mathrm{chain}}}
\DeclareMathOperator{\logit}{logit}

\begin{document}
\maketitle

\begin{abstract}
Annealed Gibbs sampling samples a multimodal model by redrawing
one variable at a time from its conditional distribution as the
temperature falls from a high value, where the distribution is broad, to the
target temperature. Each run ends in a single mode, and unbiased Monte Carlo
estimates require the runs to end in each mode in proportion to its
probability under the model. At a finite cooling rate this can fail. We
measure the error by $c$, the fraction of the model's log-odds between two
modes that the samples retain. In the Curie--Weiss model, where a weak
external field favors one of two modes, each run selects its mode near the
transition at which the modes form, and rarely switches once they have
separated. Near the transition, the random fluctuations of the Gibbs
updates still outweigh the weak field, which therefore biases the choice of
mode only slightly. Building on the selection probability of Kondepudi and coworkers,
we derive a law for $c$ that agrees with an exact computation to within
$2\%$ on nine linear cooling ramps. For the Hopfield model,
whose modes are stored patterns and their negatives, we derive a factor by
which competition among patterns raises $c$. The $c$ measured
over random pattern sets lies within $3$--$6\%$ of its prediction, but single
pattern sets depart from it, contrary to its derivation.
\end{abstract}

\begin{figure}[!ht]
 \centering
 \includegraphics[width=0.70\textwidth]{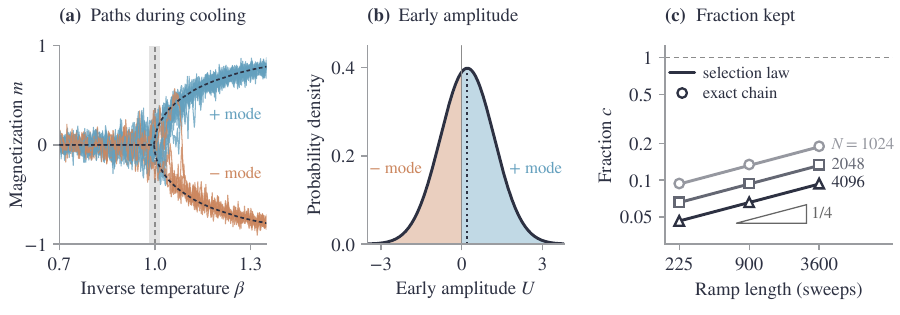}
 \vspace{-12pt}
 \caption{\textbf{Where the samples lose the model's preference.} (a) The
 magnetization $m$ of the CW model, the mean of its binary variables, cooled
 from high temperature (left) without a field (drift-and-noise approximation). 
 Runs split into the $+$ and $-$ modes at the
 transition, $\beta=1$ (band: scale of the selection window). (b) The early
 amplitude: the imbalance built up by field and noise around the transition;
 its sign decides the run's mode. A weak field toward $+$ shifts its distribution by a fifth
 of its spread: $58\%$ of runs end in $+$, which the model gives $97\%$. (c) The
 fraction $c$ of the model's log-odds kept by the samples against cooling
 time: exact (markers), our law (lines); sixteen times longer cooling only
 doubles $c$.}
 \label{fig:teaser}
 \par\vspace{1pt}
 {\centering\footnotesize\color{black!65}%
  \faGithub\enspace Code and results:\enspace
  \href{https://github.com/ali-izhar/finite-rate-selection-law}{\textcolor{black!80}{\texttt{github.com/ali-izhar/finite-rate-selection-law}}}\par}
\end{figure}

\section{Introduction}
\label{sec:intro}

Sampling from a distribution with several well-separated modes is hard,
because a Markov chain started in one mode can take a very long time to
leave it \citep{levinluczak2010,dlpcmp2009}. Annealing is meant to avoid
this: the chain starts at a high temperature, where the distribution is
broad, and is cooled to the target temperature. Each run then ends in one of
the modes. For Monte Carlo estimates to be unbiased, the runs must end in
each mode in proportion to its probability, and \citet{neal2001} noted that
there is no reason to expect annealing to achieve this. We ask how far
annealed samples miss these proportions, and why. The question
also applies to diffusion models, which commit to one class partway through
generation \citep{biroli2024}. We answer it for annealed Gibbs sampling in an
exactly solvable model.

The answer depends on when a run chooses its mode. At high temperature the
distribution has a single broad peak. The modes appear as the temperature
crosses a transition, and soon afterwards a run rarely crosses between them
in the time that remains (Figure~\ref{fig:teaser}a). The choice is therefore
made mainly near the transition. There, if the model favors one mode only
through a weak bias, this preference is still small compared with the random
fluctuations of the sampler's updates. The choice is then only slightly
better than a coin toss, even if the model at the target temperature
strongly prefers one mode (Figure~\ref{fig:teaser}b), and the samples keep
only a fraction $c$ of the model's log-odds between the two modes: with
$c=0.2$, the model's odds of four to one become
$4^{0.2}\approx1.32$ to one in the samples.

We make this quantitative for the Curie--Weiss (CW) model, in which binary
variables all interact and tend to agree, and a weak external field favors
one of its two modes. \citet{kondepudinelson1983} showed that such a choice
can be very sensitive to a small bias, and \citet{kondepudigao1987} computed
the probability that a system driven at a finite rate through a
symmetry-breaking transition ends on the favored side
\citep[see also][]{nene2018}. We turn this probability into a law for $c$
that agrees with an exact computation of the sampler's distribution to
within $2\%$ on nine linear cooling ramps. The law explains why slower
cooling helps little: for a weak field, $c$ grows only as the fourth root of
the cooling time, so cooling sixteen times more slowly only doubles $c$
(Figure~\ref{fig:teaser}c).

Many models have more than two modes, so we also ask what competition among
several modes does to $c$. In a Hopfield model the modes are stored patterns and
their negatives. If a run ends in the pattern toward which its imbalance is
largest when the modes form, competition should raise $c$: the winning
imbalance is large, and the sign of a large imbalance follows the field more
reliably than that of a small one. We derive the factor by which competition
raises $c$, assuming that the patterns of a set are statistically alike. The $c$
measured over many random pattern sets lies within $3$--$6\%$ of its
prediction, but in a post hoc analysis single pattern sets are not alike and
depart from it (Section~\ref{sec:test}). The factor therefore describes at
most an average over pattern sets.

\section{Background}
\label{sec:model}

\paragraph{Modes and their proportions.}
Many distributions of interest put most of their probability in a few
separated regions of configuration space, their \emph{modes}, with little
probability in between. For two modes A and B, let $q$ be the probability of
A given that a configuration lies in A or B. Its log-odds,
$\logit q=\log[q/(1-q)]$, are zero for equally probable modes and grow in
magnitude as one mode dominates.

\paragraph{Temperature, Gibbs sampling and annealing.}
Such distributions are often written as $\pi_\beta(x)\propto
e^{-\beta E(x)}$, with an energy $E$ and an \emph{inverse temperature}
$\beta$. At small $\beta$, high temperature, $\pi_\beta$ is broad; at large
$\beta$, low temperature, it concentrates on configurations of low energy,
where modes form. Gibbs sampling picks one
variable uniformly at random, redraws it from its conditional distribution
given all the others, and repeats. At fixed $\beta$ this Markov
chain converges to $\pi_\beta$, its \emph{equilibrium distribution}.
Convergence is slow when $\pi_\beta$ has separated modes, because the chain
must pass through improbable configurations between them: in the CW model
below, without a field and for $\beta>1$, this time grows exponentially with
the number $N$ of variables \citep{levinluczak2010,dlpcmp2009}. Annealing starts at a small $\beta$, where the
chain moves freely, and raises $\beta$ gradually. We use linear ramps, on which $\beta$ rises by the \emph{cooling
rate} $\gamma$ per \emph{sweep} of $N$ updates, one update per variable on
average. The final
configuration of a run is its sample. If $\beta$ rose
infinitely slowly at fixed $N$, the chain would stay in equilibrium and the
samples would have the equilibrium proportions. At a finite cooling rate
they need not.

\paragraph{The Curie--Weiss model.}
A configuration of the CW model is a vector $x=(x_1,\ldots,x_N)$ of $N$
binary variables $x_i\in\{-1,+1\}$, called spins. Every pair of spins interacts equally, and configurations
whose spins agree are more probable:
\begin{equation}
 \pi_{\beta,G}(x)\propto
 \exp\!\left[\frac{\beta}{2(N-1)}\sum_{i\ne j}x_i x_j
              +\beta G\sum_i x_i\right].
 \label{eq:model}
\end{equation}
The \emph{field} $G>0$ favors positive spins. We summarize a configuration by its
\emph{magnetization} $m=N^{-1}\sum_i x_i$, the mean spin. Without a field,
as $N\to\infty$, the equilibrium $m$ concentrates at $0$ for $\beta<1$ and at
two opposite values for $\beta>1$. Configurations near either value form the
two modes, which appear at the \emph{transition}, $\beta=1$. A weak field, of
order $1/N$ as in our experiments, leaves both modes in place but gives the
positive one more probability, and the log-odds between the modes grow as
$\beta$ rises. The field on each spin is small ($NG=1.25$ at every size, so
$G\approx0.0006$ at $N=2048$) but acts on all $N$ spins. At $\beta=1.6$,
where our ramps end, the equilibrium log-odds are $3.57$, so the positive
mode is $e^{3.57}\approx36$ times as probable as the negative one and holds
$97\%$ of the probability.

The sampler of the CW model can be computed exactly. In equilibrium, a
configuration's probability depends only on its number $k$ of positive
spins, so for $N=4$ the six configurations with two positive spins, such as
$++--$ and $+-+-$, are equally probable. With uniformly chosen updates, they also change
$k$ at the same rates. A run is therefore described by $k$ alone
\citep{levinluczak2010}. The \emph{exact chain} evolves the $N+1$
probabilities of $k$, instead of those of the $2^N$ configurations, and gives
the samples' exact distribution.

\paragraph{What we measure.}
Let $\qchain$ be the probability that a sample has $m>0$, and $\qeq$ the
probability of $m>0$ in equilibrium at the final $\beta$, both conditioned on
$m\ne0$, which belongs to neither mode. For $\qeq\ne1/2$, the samples keep the fraction
\begin{equation}
 c=\frac{\logit\qchain}{\logit\qeq}
 \label{eq:c}
\end{equation}
of the model's log-odds. Equivalently, the samples' odds are the model's
odds raised to the power $c$. A value $c=1$ reproduces the equilibrium
proportions, and $c=0$ gives equal proportions.

\paragraph{Several modes: the Hopfield model.}
In the Hopfield model, the coupling of Eq.~\eqref{eq:model} is built from
$p\ll N$ stored patterns $\xi^\mu\in\{-1,+1\}^N$: $x_ix_j$
becomes $\sum_\mu\xi_i^\mu\xi_j^\mu x_ix_j$. At the temperatures we use
($\beta\le1.6$), the modes are the patterns and their negatives
\citep{ags1985}, so a run chooses among $2p$ modes. The \emph{overlap}
$m^\mu=N^{-1}\sum_i\xi_i^\mu x_i$ measures how close a configuration is to
pattern $\mu$; it is $+1$ at the pattern and $-1$ at its negative. We draw
the pattern entries independently, each $+1$ or $-1$ with equal probability, and call one
draw of the $p$ patterns a \emph{pattern set}. The field acts along
pattern~1: the field term of Eq.~\eqref{eq:model} becomes $\beta
G\sum_i\xi_i^1x_i$. Pattern $\mu$ \emph{wins} a run when $|m^\mu|$ is the
largest overlap magnitude in its sample. The field favors
pattern~1 over its negative, so we define $c$ for this pair. Among the
runs that pattern~1 wins, $\qchain$ is the probability that $m^1>0$, and
$\qeq$ is its equilibrium counterpart, taken from the CW model. This $c$ concerns the sign of pattern~1, not how often
it wins.

\section{The selection law}
\label{sec:law}

We now derive the probability that a run selects the favored mode, and from
it the fraction $c$. We follow the magnetization $m(t)$ of one run, where $t$
is the time in sweeps measured from the moment $\beta$ crosses $1$. Near
this moment $m$ is small and $\beta\approx1$, and to leading order it obeys
the stochastic differential equation (Appendix~\ref{app:exact})
\begin{equation}
 dm=\big[(\beta(t)-1)\,m+G\big]\,dt+\sqrt{2/N}\;dW_t ,
 \label{eq:sde}
\end{equation}
where $W$ is a standard Brownian motion. The last term is the noise: the
random choices of the updates add a variance of about $2/N$ per sweep. On a
linear ramp, $\beta(t)-1=\gamma t$.

\paragraph{Three phases.}
The coefficient $\gamma t=\beta-1$ of $m$ in Eq.~\eqref{eq:sde} combines two
parts of the drift: $+\beta m$ from the coupling, which aligns each spin with
the majority, and $-m$ from the redraws, which reset it. It is negative before
the transition and positive after it. Without field and noise,
Eq.~\eqref{eq:sde} reduces to the ordinary differential equation
$dm/dt=\gamma t\,m$, whose solution is $m(t)=m(0)\,e^{\gamma t^2/2}$. The
exponent depends on $t^2$, which gives three phases. Before the transition
($t<0$), $t^2$ falls toward zero, so $|m|$ decreases: from a start at $t_0$,
the initial imbalance is multiplied by $e^{-\gamma t_0^2/2}$. In the
\emph{selection window}, $|t|\le T=\gamma^{-1/2}$ sweeps, the exponent is at
most $1/2$, so the drift changes $|m|$ by at most a factor $e^{1/2}$, and the sign
of $m$ is set mainly by the field and the noise. After the window ($t>T$), the
exponent grows as $t^2$, so $|m|$ grows faster than exponentially, the field and the
noise matter less and less, and the drift carries $m$ into a mode, which the run rarely leaves once the modes have separated.

\paragraph{The early amplitude.}
To find the sign that the window leaves, we divide out the growth factor,
$Y(t)=e^{-\gamma t^2/2}m(t)$. The factor is deterministic, so the product
rule applies to Eq.~\eqref{eq:sde}, and the term $\gamma t\,m\,dt$ cancels:
\[
 dY=e^{-\gamma t^2/2}\big(G\,dt+\sqrt{2/N}\,dW_t\big).
\]
Only the field and the noise change $Y$, each with the weight
$e^{-\gamma t^2/2}=e^{-t^2/2T^2}$, a Gaussian in $t$ of width $T$. $Y$
therefore settles within a few $T$ of the transition, and since
$m=e^{\gamma t^2/2}\,Y$ with a positive factor, its final value, the
\emph{early amplitude}, has the sign of the run's mode. $Y$ sums the field and
independent Gaussian noise increments, so it is Gaussian across runs; for a
ramp that starts well before the window, its mean sums the weights and its
variance sums their squares:
\[
 \mathbb E[Y]=G\!\int_{-\infty}^{\infty}\!e^{-\gamma t^2/2}dt=G\sqrt{2\pi/\gamma},
 \qquad
 \mathrm{Var}[Y]=\frac2N\!\int_{-\infty}^{\infty}\!e^{-\gamma t^2}dt=\frac2N\sqrt{\pi/\gamma}.
\]
Both grow in proportion to $T$: the mean because the field pushes in one
direction throughout the window, the variance because the noise increments
are independent. The standard deviation therefore grows only as $T^{1/2}$. In
units of it, $U=Y/\sqrt{\mathrm{Var}[Y]}\sim\mathcal N(a,1)$, and a run ends
with $m>0$ when $U>0$, which gives the classical selection probability
\citep{kondepudinelson1983,kondepudigao1987}
\begin{equation}
 q_{\mathrm{sel}}=\Phi(a),\qquad
 a=\frac{\mathbb E[Y]}{\sqrt{\mathrm{Var}[Y]}}=G\sqrt N\,(\pi/\gamma)^{1/4},
 \label{eq:law}
\end{equation}
where $\Phi$ is the standard Gaussian CDF. Since $a\propto T^{1/2}=\gamma^{-1/4}$,
a slower ramp raises $a$ only as the fourth root of its length
\citep{berglundgentz2002}.

\paragraph{The selection law.}
If each run ends with the sign of its early amplitude,
$\qchain=q_{\mathrm{sel}}$, and
Eq.~\eqref{eq:c} gives the \emph{selection law},
$c_{\mathrm{sel}}=\logit\Phi(a)/\logit\qeq$. For a weak field both log-odds
are proportional to $G$. In the numerator, $\Phi(a)\simeq\frac12+a/\sqrt{2\pi}$
for $|a|\ll1$, and $\logit q\simeq4(q-\frac12)$ near $q=\frac12$. In the
denominator, the field adds $\beta_fNGm$ to the log weight of
Eq.~\eqref{eq:model}, so the two modes, mirror images at $m=\pm\bar m_f$,
differ in log weight by $2\beta_fNG\bar m_f$ (Appendix~\ref{app:exact}):
\[
 \logit\Phi(a)\simeq\frac{4a}{\sqrt{2\pi}},
 \qquad
 \logit\qeq\simeq2\beta_fNG\bar m_f,
\]
where $\beta_f$ is the final inverse temperature and $\bar m_f\approx0.89$ at
$\beta_f=1.6$. The field cancels in their ratio:
\begin{equation}
 c\simeq\frac{\sqrt2}{\pi^{1/4}\beta_f\bar m_f}\,(N^2\gamma)^{-1/4}.
 \label{eq:scaling}
\end{equation}
For a weak field the law thus predicts that $c$ does not depend on the field,
that a sixteen times slower ramp doubles $c$, and that a fourfold larger $N$
halves it. On nine linear ramps, the exact chain of
Section~\ref{sec:model} exceeds $c_{\mathrm{sel}}$, without the
weak-field approximations, by only $0.23$--$1.81\%$
(Table~\ref{tab:linear}).

The derivation assumes a slow ramp ($\gamma\ll1$) that starts well below
$\beta=1$, a drift that $N\gamma\gg1$ keeps nearly linear in the window without a
field \citep{berglundgentz2002}, and little switching once the modes have
separated, which is far slower than our ramps. The law holds even where they
are strained (Appendix~\ref{app:exact}).

\section{Competition among several modes}
\label{sec:test}

We now ask how competition among the $2p$ modes of the Hopfield model
(Section~\ref{sec:model}) changes $c$.

\begin{figure}[!t]
 \centering
 \includegraphics[width=0.58\textwidth]{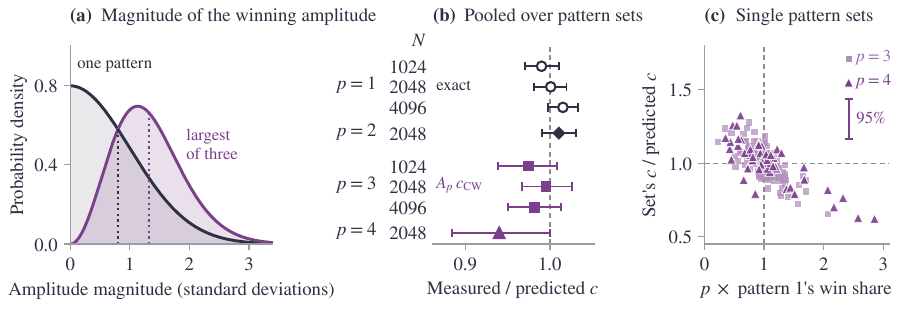}
 \caption{\textbf{Competition among patterns.} (a) Densities of the magnitude
 of one Gaussian amplitude and of the largest of three: the winner is larger
 on average, by a factor $A_3=1.662$ (dotted: means). (b) Measured over predicted $c$, pooled over
 $50$ random pattern sets, with $95\%$ intervals (dark: exact chain and two-block predictions;
 purple: $A_pc_{\mathrm{CW}}$). (c) The same ratio for each set with $p\ge3$,
 against $p$ times pattern~1's win share; alike patterns would put every set
 at $(1,1)$ (bar: median $95\%$ interval of a four-pattern set).}
 \label{fig:competition}
 \vspace{-10pt}
\end{figure}

\paragraph{Prediction.}
Each overlap $m^\mu$ follows an equation like Eq.~\eqref{eq:sde}, so each has
an early amplitude $Y^\mu$. We assume that the $Y^\mu$ are independent
Gaussians of common variance $\sigma^2$, that the field shifts only the mean
of $Y^1$, by $\delta$, and that the growth after the window carries the run
to the pattern whose amplitude is largest in magnitude, with that amplitude's
sign.
Two facts then determine $c$. First, a large amplitude follows the field more
reliably than a small one: at magnitude $r$, the odds that $Y^1$ is positive
are the ratio of its density $f$ at $+r$ and $-r$,
$f(r)/f(-r)=e^{2r\delta/\sigma^2}$, so for a weak field the probability of a
positive sign is $\frac12+r\delta/(2\sigma^2)$, linear in $r$. Second, which pattern wins depends only on the
magnitudes, and the winner's magnitude is the largest of $p$, so it is larger
on average than that of a single amplitude
(Figure~\ref{fig:competition}a). Because the probability is linear in $r$,
averaging it over the winner's magnitude rather than a single magnitude
multiplies the log-odds that $m^1$ has the field's sign, among the runs that
pattern~1 wins, by the ratio of the two mean magnitudes
(Appendix~\ref{app:amplification}):
\begin{equation}
 A_p=\frac{\mathbb E\max_{1\le\mu\le p}|Z_\mu|}{\mathbb E|Z|},
 \qquad c_{\mathrm{pred}}=A_p\,c_{\mathrm{CW}},
 \label{eq:amplification}
\end{equation}
where $Z$ and the $Z_\mu$ are independent standard Gaussians, so
$A_2=\sqrt2$, $A_3=1.662$ and $A_4=1.836$. Here $c_{\mathrm{CW}}$ is the
exact CW fraction on the same ramp; $c_{\mathrm{pred}}$ also assumes that
other patterns change neither $\delta/\sigma$ nor the equilibrium log-odds.

\paragraph{Test.}
We simulated Gibbs runs on the fastest of the nine ramps of
Section~\ref{sec:law} ($\gamma=0.004$), where $N\gamma$ is largest, with
$p=3$ at $N=1024$, $2048$ and $4096$ and $p=4$ at $N=2048$, each over $50$
random pattern sets (Appendix~\ref{app:protocol}). Runs with one pattern,
whose $c$ is known exactly, and with two, whose $c$ follows from two exactly
computed blocks (Appendix~\ref{app:amplification}), reproduced it to within
$1.6\%$ (Figure~\ref{fig:competition}b). Pooled over the pattern sets, the measured
$c$ was within $3\%$ of $A_3c_{\mathrm{CW}}$ for three patterns at all three
sizes, and $6\%$ below $A_4c_{\mathrm{CW}}$ for four (point estimates). Single pattern sets,
however, depart from the prediction (Figure~\ref{fig:competition}c): with four
patterns, pattern~1 wins between $9\%$ and $71\%$ of the runs of a set
rather than a quarter, and the $c$ of a set falls as this share rises.
Weighted equally (post hoc), the sets match the prediction on average
(measured over predicted $c$ is $0.99$--$1.01$). The factor $A_p$ thus
describes averages over pattern sets but not a single set, whose patterns are not alike as its
derivation assumes, so the test does not support the derivation.

\section{Limitations and related work}
\label{sec:limits}

Why single pattern sets differ is open. One candidate is the chance
correlation between random patterns, of order $N^{-1/2}$: it makes the
amplitudes' variances unequal unless $p/(N\gamma)\ll1$
(Appendix~\ref{app:amplification}), whereas $p/(N\gamma)$ is $0.18$--$0.73$
here, and a pattern whose amplitude varies more would win more often
and have lower sign log-odds, as in Figure~\ref{fig:competition}c.
Runs at several cooling rates could test it.
Both models are fully connected. The check of the law covers linear ramps, one value of $NG$ and $a\le0.42$, so it does
not test the law's prediction that $c$ is independent of the field. The
Hopfield test covers one cooling rate, and its ratios do not test the
equilibrium log-odds, which enter both sides.

\paragraph{Related work.}
\label{sec:discussion}%
Diffusion models with an exact score reverse a noising process rather than
cool a Gibbs chain, so our calculation does not cover them
\citep{biroli2024}; there, too few uniform steps of a variance-preserving
process get the mode probabilities of a Gaussian mixture wrong, and schedules
that slow down near the transition recover them, also for a two-mode CW
distribution \citep{aranguri2025}. Annealed importance sampling corrects
annealed estimates with importance weights \citep{neal2001}, and a cooling
study of Ising chains \citep{suzuki2009} measures defects and excess energy,
not mode probabilities.

\enlargethispage{\baselineskip}
\paragraph{Conclusion.}
In the CW model, annealed Gibbs samples keep only a fraction $c$ of the
model's log-odds. A law built on the selection of each run's mode near the
transition, where the random fluctuations of the updates outweigh a weak
field, gives $c$ to within $2\%$ on nine linear ramps. With several modes, the
competition factor matches averages over pattern sets but not single sets.
A theory for single sets and the extension to diffusion models remain open.

\label{main-text-end}
\bibliography{references}

\appendix

\section{Derivation of the selection law}
\label{app:exact}

\paragraph{Drift and noise of $m$.}
By Eq.~\eqref{eq:model}, a Gibbs update redraws spin $i$ with mean
$\tanh\ell_i$, where $\ell_i=\beta(N-1)^{-1}\sum_{j\ne i}x_j+\beta G$. The sum
differs from $m$ by terms of order $1/N$, so for large $N$ this mean is
$\tanh(\beta m+\beta G)$. An update changes $m$ by the new value minus the old
value of the chosen spin, divided by $N$. The spin is chosen uniformly, so its
old value has mean $m$, and the $N$ updates of a sweep give
\[
 \dot m=-m+\tanh(\beta m+\beta G).
\]
Expanding $\tanh u=u-u^3/3+\dots$ for small $m$ and $G$ gives
$(\beta-1)m+\beta G-\beta^3m^3/3$; near $\beta=1$, $\beta G\simeq G$, and
$-m^3/3$ is the leading term that Eq.~\eqref{eq:sde} omits. Near $m=0$, the
redrawn value is $\pm1$ with probability about one half each, independently
of the old value, so an update flips its spin with probability about one half,
and a flip changes $m$ by $2/N$. The $N$ updates of a sweep therefore add a
variance of about $N\times\tfrac12\times(2/N)^2=2/N$, which we treat as
Gaussian noise.

\paragraph{Start of the ramp.}
From a start at $t_0<0$, $Y$ starts at $Y(t_0)=e^{-\gamma t_0^2/2}m(t_0)$, so
the initial imbalance enters the early amplitude with weight
$e^{-\gamma t_0^2/2}$. Our ramps start at $\beta=0.7$, that is,
$t_0=-0.3/\gamma$, so $\gamma t_0^2/2=0.045/\gamma\ge11$ and this weight is
below $e^{-11}$; the integrals of Section~\ref{sec:law} can then run over the
whole line.

\paragraph{Weak-field log-odds.}
Let $K(s)=\log\mathbb E_+[e^{sm}]$ be the cumulant generating function of
$m$ in the positive mode at zero field. The negative mode is its mirror
image, so with $s=\beta_fNG$, $\logit\qeq=K(s)-K(-s)$. The even cumulants
cancel, and
\[
 \logit\qeq=2\big(\kappa_1s+\kappa_3s^3/6+\dots\big),
\]
where $\kappa_1=\bar m_f$ is the mean magnetization of the positive mode,
close to the positive solution of $m=\tanh(\beta_fm)$, and $\kappa_3$, the
third cumulant of a mean of $N$ spins, is of order $1/N^2$. Since $s$ is of
order one, the correction to the leading term of Section~\ref{sec:law} is
smaller by a factor of order $1/N^2$.

\paragraph{When the law holds.}
The derivation needs the drift to stay linear until the sign is set. At the
edge of the window, $\gamma t=\sqrt\gamma$, and the omitted cubic term
$m^3/3$ matches the linear term $\sqrt\gamma\,m$ once $m^2\sim\sqrt\gamma$,
that is, $|m|\sim\gamma^{1/4}$. There, by Section~\ref{sec:law}, the noise
gives $m$ a standard deviation of order $(T/N)^{1/2}=N^{-1/2}\gamma^{-1/4}$,
a fraction $(N\gamma)^{-1/2}$ of this scale, and for $a$ of order one the
field's contribution is of the same order. So $N\gamma\gg1$ keeps the drift
nearly linear up to the edge of the window; \citet{berglundgentz2002} prove
this for the pitchfork without a field. Since $\beta-1=\gamma t$, the window
$|t|\le T$ is $|\beta-1|\le\sqrt\gamma$, and the same condition makes it
wider than the finite-size critical window $|\beta-1|\sim N^{-1/2}$.

\paragraph{Where the assumptions are strained.}
The law holds even where these conditions are strained, for reasons we did
not derive: $N\gamma$ ranges from $0.26$ to $16$ on the nine ramps,
and on the slowest ramp at $N=1024$, $12\%$ of runs end with the sign of $m$
opposite to its sign at $\beta=1.05$, just after $t=3T$, yet the law matches the final $c$
to $1\%$. Switching would restore equilibrium on a much slower ramp, but at
$N=2048$ and $\beta=1.1$ a zero-field switch takes about $5\times10^{7}$
sweeps ($5\times10^{4}$ at $N=1024$), while our ramps last at most $3600$.
These times are mean passage times of the exact chain between the centers of
the two modes, at fixed $\beta$ and zero field.

\paragraph{Exact chain.}
Updates arrive at rate $N$ per sweep, so the number $k$ of positive
spins is a continuous-time Markov chain. It rises or falls by one at a rate
equal to the number of eligible spins ($N-k$ negative or $k$ positive) times
their redraw probability:
\[
 W_k^+=\frac{N-k}{2}\Big[1+\tanh\!\Big(\beta\frac{2k-N+1}{N-1}+\beta G\Big)\Big],
 \qquad
 W_k^-=\frac{k}{2}\Big[1-\tanh\!\Big(\beta\frac{2k-N-1}{N-1}+\beta G\Big)\Big],
\]
where the $\pm1$ exclude the updated spin. We integrate the probabilities
$P_k(t)$ of the $N+1$ counts along the ramp with these rates, which gives
$\qchain$; the integrator takes implicit Euler steps of about $0.02$ sweeps
with Richardson extrapolation and holds $\beta$ at the midpoint of each
interval $\Delta\beta=0.005$ for $0.005/\gamma$ sweeps. With $\pi_k$ the
equilibrium distribution of $k$ at the final $\beta$,
$\qeq=\sum_{k>N/2}\pi_k/\sum_{k\ne N/2}\pi_k$. The excluded state $m=0$ has
equilibrium probability below $10^{-36}$ at $\beta_f=1.6$ for $N\ge512$.

\paragraph{The nine ramps.}
The ramps of Table~\ref{tab:linear} run from equilibrium at $\beta=0.7$ to
$\beta=1.6$, with $G=0.0006125\times2048/N$, that is, $NG=1.2544$. A fixed
$NG$ keeps the equilibrium log-odds nearly constant across sizes. At each $N$,
runs whose final sign differs from their sign just after $t=3T$ are most
frequent on the slowest ramp; at $N=2048$, this occurs in $3\%$ of runs. We
take that sign at $\beta=1.05$, the first grid point at or above
$1+3\sqrt\gamma$. These percentages do not count runs that cross and return. The discrete-time
chain of the Hopfield test (Table~\ref{tab:settings}), which needs no time
step, reproduces the $\gamma=0.004$ column to the printed digits.

\paragraph{Settings of Figure~\ref{fig:teaser}.}
(a) Fourteen paths of a drift-and-noise approximation of the sampler, with
drift $-m+\tanh(\beta m)$ and the noise of Eq.~\eqref{eq:sde}, at $N=1024$,
zero field and $\gamma\approx0.0003$; the band marks $|\beta-1|<\sqrt\gamma$. (b) $N=2048$, $\gamma=0.001$, $NG=1.2544$, so $a=0.208$. (c) The nine
ramps of Table~\ref{tab:linear}, whose length is $0.9/\gamma$ sweeps.

\begin{table}[!htb]
 \centering
 \caption{Exact $c$ on the nine linear ramps of Section~\ref{sec:law} and, in
 parentheses, its percentage excess over the selection law $c_{\mathrm{sel}}$.}
 \label{tab:linear}
 \begin{tabular}{rrrr}
 \toprule
 $N$ & $\gamma=0.004$ & $\gamma=0.001$ & $\gamma=0.00025$\\
 \midrule
 1024 & $0.09370\ (0.93\%)$ & $0.13393\ (1.81\%)$ & $0.18868\ (1.03\%)$\\
 2048 & $0.06587\ (0.46\%)$ & $0.09372\ (0.97\%)$ & $0.13314\ (1.23\%)$\\
 4096 & $0.04645\ (0.23\%)$ & $0.06589\ (0.49\%)$ & $0.09361\ (0.86\%)$\\
 \bottomrule
 \end{tabular}
\end{table}

\section{The competition factor}
\label{app:amplification}

\paragraph{Derivation of $A_p$.}
Section~\ref{sec:test} gives the probability
$\frac12+r\delta/(2\sigma^2)$ that $Y^1$ is positive at magnitude $r$. The
winner depends only on the magnitudes, so this probability also holds among
the runs that pattern~1 wins. It remains to find the distribution of the
winner's magnitude. Reversing the field reverses the sign of $Y^1$ and leaves
the magnitudes unchanged in distribution, so this distribution is an even
function of $\delta$. An even function has no first-order term, so to first
order in $\delta$ the distribution is the one at $\delta=0$, where the
patterns are alike: that of $\sigma M_p$, with $M_p=\max_{1\le\mu\le p}|Z_\mu|$.
Averaging the sign probability over it, and using $\logit q\simeq4(q-\frac12)$,
gives
\[
 \qchain\simeq\frac12+\frac{\delta}{2\sigma}\,\mathbb E M_p,
 \qquad
 \logit\qchain\simeq\frac{2\delta}{\sigma}\,\mathbb E M_p,
 \qquad
 \mathbb E M_p=\int_0^\infty\big[1-(2\Phi(x)-1)^p\big]\,dx .
\]
The last form is $\mathbb EM_p=\int_0^\infty\Pr(M_p>x)\,dx$, since $M_p\le x$
exactly when all $p$ magnitudes are at most $x$. With one pattern, $M_1=|Z|$
and this is the weak-field result of Section~\ref{sec:law}. The ratio of the two log-odds is $A_p$.

\paragraph{One pattern.}
With $y_i=\xi_i^1x_i$, the Hopfield update of $x_i$ is the CW update of
$y_i$. The updated spin is chosen independently of the state, so the
one-pattern sampler is the CW sampler for every $N$ and every pattern, and
its $c$ is the exact-chain value.

\paragraph{Two patterns.}
In the variables $y_i$, the coupling of $y_i$ and $y_j$ is
$(1+\eta_i\eta_j)/(N-1)$, with $\eta_i=\xi_i^1\xi_i^2$. It vanishes between
the spins on which the two patterns agree and those on which they differ, and
equals $2/(N-1)$ within each group. The model therefore consists of two
non-interacting CW blocks ($944$--$1104$ spins in our pattern sets), each with
the field $G$; a block of $n$ spins has its transition at $\beta\simeq N/(2n)$,
within $0.09$ of $\beta=1$. With $S_A$ and $S_B$ the block sums of the $y_i$,
$Nm^1=S_A+S_B$ and $Nm^2=S_A-S_B$, so pattern~1 wins exactly when $S_A$ and
$S_B$ have the same sign, which is then the sign of $m^1$. We compute each
block's distribution exactly and treat the blocks as independent, which they
are up to the shared number of updates. The odds that both sums are positive
rather than both negative are then the product of the blocks' odds, and with
pooling over pattern sets as the estimator does, this gives
$c_{\mathrm{pred}}=0.09363$ at $N=2048$. By Eq.~\eqref{eq:law}, the sampled
log-odds of a block grow as the square root of its size, and the log-odds of the
two blocks add, so two blocks of about $N/2$ spins give
$2\sqrt{1/2}=\sqrt2$ times those of one block of $N$. This recovers
$A_2=\sqrt2$, and $A_2c_{\mathrm{CW}}=0.09316$ lies $0.5\%$ below the
two-block value.

\paragraph{When random patterns violate the assumptions.}
Let $C_{\mu\nu}=N^{-1}\sum_i\xi_i^\mu\xi_i^\nu$ be the correlation of
patterns $\mu$ and $\nu$. At zero field and for small overlaps, averaging the
expected Gibbs redraw over spins, weighted by $\xi_i^\mu$, gives to leading
order in $1/N$
\[
 \dot m^\mu\simeq-m^\mu+\beta\sum_\nu C_{\mu\nu}m^\nu .
\]
The component of $(m^1,\ldots,m^p)$ along an eigenvector of $C$ with
eigenvalue $\lambda$ therefore begins to grow at $\beta=1/\lambda$ and not at
$\beta=1$. An off-diagonal entry of $C$ averages $N$ independent signs and
has standard deviation $N^{-1/2}$, and the eigenvalues of a $p\times p$
symmetric matrix with independent entries of this size spread over a range of
order $(p/N)^{1/2}$. The selection window spans a $\beta$ range of order
$\sqrt\gamma$. As an order-of-magnitude condition, the components start to grow
at nearly the same point of the window only when
\[
 (p/N)^{1/2}\ll\sqrt\gamma,\qquad\text{that is,}\qquad p/(N\gamma)\ll1 .
\]
Otherwise the components grow by unequal factors, and since each overlap
mixes the components, the amplitudes of the patterns can then have unequal
variances.

\section{The Hopfield test}
\label{app:protocol}

Tables~\ref{tab:settings} and~\ref{tab:transfer} give the settings and the
results of the test of Section~\ref{sec:test}.

\begin{table}[!htb]
 \centering
 \caption{Settings of the Hopfield test of Section~\ref{sec:test}.}
 \label{tab:settings}
 \small
 \begin{tabular}{@{}lp{0.74\textwidth}@{}}
 \toprule
 Ramp & Update $n$ of $M=225N$ uses $\beta_n=0.7+0.9(n+\tfrac12)/M$, so the ramp lasts $225$ sweeps of $N$ updates ($\gamma=0.004$).\\
 Update & Redraws one uniformly chosen spin.\\
 Start & Independent unbiased spins, the equilibrium at $\beta=0$.\\
 Field & $NG=1.2544$ along pattern~1, half of the runs with each sign; $a=0.21$, $0.15$ and $0.10$ at $N=1024$, $2048$ and $4096$.\\
 Pattern sets & $50$ for each $p$ and $N$, with equal numbers of runs. A set is the first $p$ of four vectors of independent equiprobable signs, so the sets are nested in $p$; set $s$ comes from NumPy \texttt{default\_rng}($[20261027,N,s]$).\\
 Winner & The pattern with the largest $|m^\mu|$ at $\beta=1.6$. No run ended in a tie.\\
 Estimator & $\widehat c=\logit\widehat q/\logit\qeq^{\mathrm{CW}}(N)$, where $\widehat q$ is the fraction of the runs won by pattern~1 whose $m^1$ has the field's sign, pooled over pattern sets and field signs.\\
 Denominator & Exact CW equilibrium log-odds at $\beta=1.6$: $3.5742$, $3.5747$ and $3.5749$ at $N=1024$, $2048$ and $4096$.\\
 Intervals & Percentiles of $10{,}000$ cluster-bootstrap resamples of the $50$ pattern sets (seed $7$).\\
 $c_{\mathrm{CW}}$ & Exact for this discrete sampler: the distribution of $k$ is multiplied by the transition matrix of each update. The blocks of the two-pattern prediction are computed in the same way.\\
 \bottomrule
 \end{tabular}
\end{table}

\paragraph{Checks on the settings.}
For the exact CW chain at $N=2048$, the independent-spin start of
Table~\ref{tab:settings} and the equilibrium at $\beta=0.7$ give the same $c$
to six digits. The CW denominator equals the Hopfield equilibrium log-odds at
$p=1$, and is within $1.3\%$ of them by the exact two-block calculation at
$p=2$ ($N=2048$) and by exact enumeration at $p=3$ ($N=512$ and $1024$); we
did not compute the Hopfield value at $p=4$. The same denominator enters
$c_{\mathrm{pred}}$, so it cancels in the ratios of Table~\ref{tab:transfer},
which compare sampled log-odds only. Each pattern set with $p\ge3$ received
$10{,}800$ to $29{,}200$ runs, half with each field sign.

\begin{table}[!htb]
 \centering
 \caption{Results of the Hopfield test. Ratio: measured over predicted $c$. Won: runs won by
 pattern~1, which alone enter the estimate.}
 \label{tab:transfer}
 \small
 \begin{tabular}{@{}rrlrlrrr@{}}
 \toprule
 $p$ & $N$ & prediction & $c_{\mathrm{pred}}$ & measured $c$ [$95\%$ interval] & ratio & runs & won\\
 \midrule
 1 & 1024 & exact chain & 0.0937 & $0.0928\ [0.0909,0.0947]$ & 0.990 & 320{,}000 & 320{,}000\\
 1 & 2048 & exact chain & 0.0659 & $0.0659\ [0.0647,0.0672]$ & 1.001 & 640{,}000 & 640{,}000\\
 1 & 4096 & exact chain & 0.0464 & $0.0472\ [0.0464,0.0480]$ & 1.015 & 1{,}260{,}000 & 1{,}260{,}000\\
 2 & 2048 & two-block & 0.0936 & $0.0946\ [0.0928,0.0965]$ & 1.011 & 640{,}000 & 321{,}598\\
 3 & 1024 & $A_3c_{\mathrm{CW}}$ & 0.1558 & $0.1518\ [0.1463,0.1571]$ & 0.975 & 540{,}000 & 181{,}324\\
 3 & 2048 & $A_3c_{\mathrm{CW}}$ & 0.1095 & $0.1090\ [0.1059,0.1123]$ & 0.995 & 820{,}000 & 274{,}222\\
 3 & 4096 & $A_3c_{\mathrm{CW}}$ & 0.0772 & $0.0758\ [0.0734,0.0782]$ & 0.982 & 1{,}460{,}000 & 504{,}568\\
 4 & 2048 & $A_4c_{\mathrm{CW}}$ & 0.1209 & $0.1137\ [0.1069,0.1210]$ & 0.940 & 940{,}000 & 254{,}495\\
 \bottomrule
 \end{tabular}
\end{table}

\paragraph{Results.}
For three patterns, the pooled $c$ falls as $N^{-0.50}$ (interval
$[0.47,0.53]$), as Eq.~\eqref{eq:scaling} predicts at a fixed rate; the exact
CW chain gives $0.506$. The exponent is
$\tfrac12\log_2[\widehat c(1024)/\widehat c(4096)]$, and its interval comes
from independent bootstraps of the two sizes. The three-pattern intervals
contain the prediction but allow deficits of up to $6\%$, and the
four-pattern ratio interval is $[0.884,1.00]$.

\paragraph{The analysis plan.}
Before the runs, we wrote down the predictions, the estimator, the run counts
and three readings of the four ratios $c/(A_pc_{\mathrm{CW}})$ with $p\ge3$.
(a) If every $95\%$ interval contains $1$, $A_p$ holds to within $3\%$, the
planned half-width. (b) If some intervals lie below $1$ and the deficits grow
with $p/(N\gamma)$, $A_p$ is the limit $N\gamma\to\infty$. (c) If some
interval lies above $1$, or the deficits are not ordered by $p/(N\gamma)$, an
assumption other than equal variance fails, and $A_p$ is not supported as
derived.

The outcome lies between (a) and (c). At the planned bootstrap seed every
interval contains $1$, but the four-pattern interval only just does: it ends
at $1.0005$, and below $1$ at $139$ of the $200$ bootstrap seeds $0$--$199$.
The point deficits, $2.5\%$, $0.5\%$ and $1.8\%$ for three patterns at
increasing $N$ and $6\%$ for four, are not ordered by $p/(N\gamma)$, which is
largest at $p=3$, $N=1024$ (rank correlation $0.6$), although the four
intervals overlap. We report reading (c), the less favorable one. One post hoc
comparison does follow the order: the rows with $p=3$ and $p=4$ at $N=2048$
share their pattern sets, and on these sets the four-pattern ratio is $0.055$
lower (paired bootstrap interval $[0.008,0.099]$, above zero at all $200$
seeds).

The plan has limits of its own. It is in the repository (\texttt{docs/}) with
all seeds, the pilot and the results, so no external record dates it before
the runs. A pilot with a tenth of the planned runs, never pooled with the
results, failed both of its one-pattern checks at $95\%$; four replicates
with new seeds and more runs passed (ratios $0.970$--$1.016$). Before any
reported run, we doubled all run counts and chose to read the four controls
at a joint $95\%$ level; their separate $95\%$ intervals also contain the
exact values. Three of the four ratios with $p\ge3$ missed the planned
half-width of $3\%$, because the sets varied more than assumed.

\begin{table}[!htb]
 \centering
 \caption{Single pattern sets ($50$ per row, $p\ge3$; post hoc). Win share: range over
 sets of the share of runs that pattern~1 won. $P$: chi-square test that all sets share one
 $\widehat q$. Spearman: rank correlation of the per-set $\widehat c_s$ with the win share.
 Excess SD: standard deviation of $\widehat c_s/c_{\mathrm{pred}}$ beyond sampling error.
 Equal weight: mean of $\widehat c_s/c_{\mathrm{pred}}$ over sets.}
 \label{tab:sets}
 \begin{tabular}{rrcrrrr}
 \toprule
 $p$ & $N$ & win share & $P$ & Spearman & excess SD & equal weight\\
 \midrule
 3 & 1024 & $0.075$--$0.692$ & $2\times10^{-12}$ & $-0.75$ & $8.0\%$ & 1.003\\
 3 & 2048 & $0.168$--$0.566$ & $2\times10^{-6}$ & $-0.69$ & $7.7\%$ & 1.013\\
 3 & 4096 & $0.229$--$0.522$ & $2\times10^{-5}$ & $-0.56$ & $6.6\%$ & 0.993\\
 4 & 2048 & $0.087$--$0.713$ & $5\times10^{-40}$ & $-0.72$ & $13.0\%$ & 1.002\\
 \bottomrule
 \end{tabular}
\end{table}

\paragraph{Single pattern sets (post hoc).}
The plan did not include single pattern sets. For each set, with field signs
pooled, we compute the share of runs that pattern~1 wins and $\widehat c_s$,
the estimator applied to that set alone (Table~\ref{tab:sets}). The win
shares show that the patterns of a set are not alike, a failure that the plan
did not list. The pooled $\widehat c$ weights each set by the runs that
pattern~1 won, so it weights most heavily the sets with a large win share and
a low $\widehat c_s$, which is why it lies below the equal-weight mean. For
the controls, the same chi-square test finds no excess spread
($P=0.56$--$0.94$), so the spread for $p\ge3$ does not come from the sampler
or the estimator.

\end{document}